# Generalizable Machine Learning Models for Predicting Data Center Server Power, Efficiency, and Throughput

Nuoa Lei[1], Arman Shehabi[1], Jun Lu[2], Zhi Cao[3], Jonathan Koomey[4], Sarah Smith[1], Eric Masanet[1,5]

[1] Energy Technologies Area, Lawrence Berkeley National Laboratory, Berkeley, CA, USA
[2] Division of Epidemiology and Biostatistics, University of Illinois, Chicago, IL, USA
[3] College of Environmental Science and Engineering, Nankai University, 38 Tongyan Road, Jinnan District, Tianjin, 300350, China
[4] Koomey Analytics, Bay Area, CA, USA
[5] Bren School of Environmental Science and Management, University of California Santa Barbara, Santa Barbara, CA, USA

## Highlights

- Server models: power, performance-to-power ratio, maximum throughput.
- Models based on configuration, workload level, and hardware availability date.
- Accurate and generalizable machine learning based server models.
- Analysis of important predictive factors.
- Investigated reliability in modeling prospective servers.

## Abstract

In the rapidly evolving digital era, comprehending the intricate dynamics influencing server power consumption, efficiency, and performance is crucial for sustainable data center operations. However, existing models lack the ability to provide a detailed and reliable understanding of these intricate relationships. This study employs a machine learning-based approach, using the SPECPower_ssj2008 database, to facilitate user-friendly and generalizable server modeling. The resulting models demonstrate high accuracy, with errors falling within approximately 10% on the testing dataset, showcasing their practical utility and generalizability. Through meticulous analysis, predictive features related to hardware availability date, server workload level, and specifications are identified, providing insights into optimizing energy conservation, efficiency, and performance in server deployment and operation. By systematically measuring biases and uncertainties, the study underscores the need for caution when employing historical data for prospective server modeling, considering the dynamic nature of technology landscapes. Collectively, this work offers valuable insights into the sustainable deployment and operation of servers in data centers, paving the way for enhanced resource use efficiency and more environmentally conscious practices.

## Keywords

## 1. Introduction

Data centers are foundational to the sustenance of contemporary digital life, facilitating an array of digital services and supporting the burgeoning digital economy. The exponential growth in artificial intelligence and data-driven applications has precipitated a marked surge in the data processing, storage, and communication services provided by these centers [1–3]. This upward trajectory is poised to persist, raising concerns about the escalating consumption of resources and its consequential environmental implications [4–8].

The resource consumption of computing data centers encompasses the energy used by servers, external storage devices, network equipment, and supporting infrastructure, as well as water used in cooling systems and the generation of offsite electricity generation [4,9]. Within this context, the electricity consumption of servers emerges as a pivotal factor influencing the sustainability of these facilities. Servers, integral to data centers, play a vital role in providing essential data, resources, and computing services. These power-hungry machines consistently draw substantial electrical power to uphold reliable data center operations [10]. Crucially, the electricity consumed by servers is converted into heat, resulting in a notable overhead in resource consumption, covering both electricity and water, for data center cooling systems and offsite electricity generation [9]. In 2018, global computing data centers were estimated to consume approximately 205 $TWh$ of electricity, with servers accounting for 103 $TWh$ of this consumption and an additional 61 $TWh$ dedicated to cooling these servers and covering power distribution loss [1]. The U.S. computing data centers in 2014 reported an electricity consumption of around 71 $TWh$ and a water usage of 639 billion liters [11], with servers directly consuming about 30 $TWh$ of electricity and cooling systems using an additional 22 $TWh$, alongside about 552 billion liters of water for cooling and electricity generation purposes [11]. These figures underscore the significant resource footprint attributable to servers within computing data center operations.

To address the escalating challenges of resource efficiency and sustainability in data centers, professionals, including operators, engineers, energy analysts, and policymakers, rely on modeling tools. These tools are designed to gauge resource consumption accurately and make informed decisions on efficiency-enhancing technologies, strategies, and policies [11–13]. Numerous prior studies have proficiently and reliably modeled the energy and water usage of data center infrastructure [9,14–17], delivering nuanced insights into the intricate connections between resource consumption and factors such as cooling system technologies, free cooling strategies, climate conditions, equipment efficiencies, and operational settings. These insights are invaluable for advancing sustainable data center operations. Similarly, extensive research efforts have been dedicated to developing server models for tasks such as load prediction [18,19], cooling system sizing [20,21], virtual power metering [22,23], dynamic voltage and frequency scaling-based energy conservation [24–27], thermal-aware data center design and control [21,27–29], and workload optimization [30–32], underscoring the utility of employing server models to improve data center energy efficiency.

Despite the importance of servers and numerous modeling efforts, a notable gap remains in understanding and modeling server power usage and efficiency (refer to section 2 for details). Criticisms from recent review papers highlight the lack of user-friendliness and generalizability in publicly available server models [33,34]. Some models, while accurate, are limited to a single or small subset of servers [19,21,22,24–26,28,30,32,35] or rely on hard-to-acquire operational (e.g., run-time signals from server operations) or circuit-level data [18,27,36–38]. Other models offer wider applicability at the expense of accuracy due to oversimplified assumptions [20]. Furthermore, our updated literature review indicates an overemphasis on power usage modeling, leaving the modeling of server power efficiency and performance relatively unexplored. Moreover, earlier models fall short in depicting variations in server power and efficiency associated with diverse configurations and temporal effects. The limitations inherent in existing server models impede decision-makers from tailoring general and systematic approaches to accommodate the anticipated growth in data center resource demand, as these models lack the ability to provide a detailed and reliable understanding of the intricate relationships between server power consumption, efficiency, and performance, considering a spectrum of technological, operational, and temporal factors.

This paper addresses these knowledge gaps by introducing a machine learning-based server modeling approach, leveraging the SPECPower_ssj2008 database [39]. Our method demonstrates exceptional accuracy and generalizability in predicting server power consumption, efficiency (i.e., performance-to-power ratio), and maximum throughput, without sacrificing user-friendliness. Through comprehensive analysis, we explore how technological, operational, and temporal factors influence server performance metrics. We conclude with a cautionary note on prospective server modeling and suggest avenues for future research in this domain.

## 2. Previous work

This section offers an overview of previous server models, outlining their key modeling approaches and objectives, and critically analyzing their strengths and weaknesses.

### 2.1. Complementary metal-oxide semiconductor circuit power model

Early research endeavors focused on estimating server power usage utilizing the complementary metal-oxide semiconductor (CMOS) circuit power model [33,37,38], a well-established method for gauging the power consumption of digital circuits. This model delineates dynamic and static power as the primary components of power consumption in digital circuits, elucidating the relationship between power consumption and circuit parameters such as transistor count, circuit current, supply voltage, and switching capacitance.

While effective at the circuit level, the CMOS circuit power model faces notable challenges when applied to estimating the power consumption of server systems. A key challenge is its requirement for intricate, often proprietary, circuit-level technological details, which are typically unavailable in the public domain. More critically, the validity and accuracy of the CMOS circuit power model are compromised at the server level, where higher-level system abstraction complicates the behavior of server power usage [33].

These challenges inherent in the CMOS circuit power model have spurred the evolution of two main streams of modeling approaches: data-driven approaches and hybrid modeling approaches.

### 2.2. Data-driven server power models

Data-driven approaches derive server models solely from experimental data or hardware specifications. Several researchers have employed both linear and non-linear regression analyses to capture the relationship between server power usage and CPU utilization [22–24,32,40,41]. A prominent study by Horvath et al. [24] introduced a linear power model achieving an R-squared value of 98%, using CPU utilization and clock frequency as parameters to explore how dynamic voltage scaling could optimize server energy consumption. Alternatively, Dhiman et al. [18] demonstrated the effectiveness of a classification approach for server power prediction, utilizing a Gaussian Mixture Model based on instructions per cycle, memory accesses per cycle, cache transitions per cycle, and CPU utilization, achieving an average prediction error of less than 10%.

While these models can achieve high accuracy in estimating server power use, their applicability is confined to a specific server or a narrow selection of servers they were trained on, necessitating retraining for accuracy across different server types. In an effort to improve the generalizability of server power prediction models, Cheung et al. [20] incorporated server-specific information (such as rated processor speed, number of processors, and the maximum power output of the power supply unit (PSU)) into a linear regression model. This model predicted idle and maximum power across various servers, which was subsequently extrapolated to estimate server power consumption at different CPU utilization levels, assuming ideal power proportionality. However, Cheung's model has limitations, notably that the maximum power output of the PSU is an outcome of server design rather than a predictive input for server design, and the model's accuracy (mean absolute percentage error of 25.7%) raises concerns for practical use. Additionally, the assumption of perfect power proportionality may not accurately reflect the power behavior of diverse servers.

### 2.3. Hybrid server power models

Beyond data-driven approaches, hybrid modeling approaches have gained prominence in the realm of server power modeling. In a hybrid power model, the server's power consumption is typically divided into multiple additive components, each representing the power consumption of a major server component. This approach blends experimental data with theoretical insights, often drawing on principles from the CMOS circuit power model. When certain server components are not considered in the power modeling, a bias term is used to compensate for potential discrepancies in power prediction. Hybrid modeling approaches offer the advantage of allowing modelers to incorporate specific parameters of interest from physics-based knowledge. For example, researchers have modeled server power usage by focusing on the dynamic power usage of the CPU [25,30]. They imposed a cubic relationship between CPU power and operating frequency and incorporated a bias term to compensate for power usage in other server components. Other studies also modeled server power consumption by considering the CPU and server fan power use [19,21,28], utilizing insights from the CMOS circuit power model and fan curve. This leads to server models that can facilitate thermal-aware data center design. Researchers like Xu et al. [26] conducted power modeling by examining the power consumption of CPUs and disks, considering the impact of CPU frequencies, CPU utilization, and disk utilization. This resulted in a framework that enables power conservation in data centers through dynamic voltage and frequency scaling. Similarly, Ge et al. [31] developed a power model for multicore servers by focusing on CPU and memory power consumption, which concurrently accounts for the impact of workload features, parallel processing across nodes and cores, and processor frequency scaling. More recently, Arroba et al. [27] focused on modeling the power usage of the CPU, memory, and server fan, leading to a server model supporting both thermal-aware data center design and energy savings through dynamic voltage and frequency scaling. Notably, the accuracy of the hybrid model can potentially be improved by incorporating additional server components into the modeling scope. For example, Basmadjian et al. [36] decomposed server power usage into six additive components (CPU power, memory power, hard disk power, network interface card power, fan power, and power loss due to the PSU), resulting in a power model with an error of 2-10%. In a similar vein, Song et al. [35] considered various server components such as CPU, memory, disk, network interface card, and network peripherals, achieving an average error of 3.5% when predicting server power usage.

While hybrid modeling approaches rival the effectiveness of data-driven techniques with lesser data requirements, they are not without limitations. First, they typically assume that server power is the linear sum of component power, overlooking the complex interplay between components. Second, during the modeling stage, restrictive parametric relationships informed from physical knowledge are often imposed, which may not accurately represent the ground truth due to oversimplification or changes in applicable conditions. Lastly, akin to previous data-driven models, the precision of hybrid approaches is confined to the specific servers they are calibrated for, limiting their broader applicability in server configuration optimization or early-phase data center power demand forecasting.

## 2.4. Performance and power-performance models

In contrast to the considerable focus on server power modeling, only a limited number of studies have delved into modeling server performance or power efficiency (i.e., power-performance ratio). Lin et al. [32] developed models using regression analysis for server performance and power efficiency as functions of CPU utilization, reporting average errors of 0.32% for performance and 3.28% for power efficiency. Ge et al. [31] utilized Amdahl's law and employed execution time as a performance metric to analyze the speedup achieved by multicore parallel processing. Song et al. [35] also utilized workload execution time as a performance metric, breaking it down into additive components, including computation time, main memory access time, storage access time, and communication overhead, with each component modeled as a regression function of problem size, parallelism, and hardware frequency.

While these models provide valuable insights for enhancing power efficiency and optimizing workloads in data centers, they face several important challenges. First, similar to server power modeling, previous performance and efficiency modeling approaches are restrictive to a small set of servers, and different model parameters need to be fitted for different servers to maintain model accuracy. Second, previous models do not offer the capability to understand how server throughput and power-performance may vary across diverse server configurations, nor have they been tested for servers equipped with newer generations of technology. Third, previous performance metrics that rely solely on execution time fail to offer insights into the maximum throughput of servers and do not provide information on energy consumption on a per-workload basis.

## 2.5. Proposed server models

This study focuses on three primary modeling objectives: server power consumption ($P_L$, $W$), maximum throughput ($Th_{max}$, $\frac{ssj\_ops}{s}$), and performance-to-power ratio ($Perf_L$, $\frac{ssj\_ops}{W}$). In our settings, server power consumption refers to the power utilized by servers at a specific workload level ($L$, %). Maximum throughput quantifies the number of server workload[1] operations per second when the server operates at full capacity. The performance-to-power ratio is computed as the server's throughput divided by its power consumption at a given workload level, serving as a metric for the server's power efficiency [39]. The relationship between $P_L$, $Perf_L$, and $Th_{max}$ can be expressed by the Eq. (1) below:

$$Perf_L = \frac{L \times Th_{max}}{P_L} \qquad (1)$$

Our approach distinguishes itself from previous studies in several aspects. Firstly, our study extends the scope of server modeling by simultaneously focusing on power consumption, maximum throughput, and performance-to-power ratio. We identify the intricate relationships of these model targets with server configuration, workload, and hardware availability year, which can be useful for data center design and optimization. Specifically, our modeling approach can reliably predict power consumption and performance-to-power ratio across various server workloads, providing insights into server power proportionality [42] and efficiencies at different workloads. Secondly, our server models, which leverage robust and reliable machine learning algorithms, outperform existing models in terms of generalizability. This is evident in their high accuracy on both observed and unobserved servers, a notable improvement compared to existing models that are confined to a single server or a small subset once trained. Finally, inputs to our server models can be easily and publicly obtained. Our approach does not need run-time measurements from sensors, nor does it require circuit-level or detailed technology information to make predictions, greatly improving the applicability of the models.

---

[1] The workload is a standardized Java program designed to assess the performance of the CPU(s), caches, memory, shared memory processors, Java Virtual Machine implementations, Just In Time compilers, garbage collection, threads, and specific aspects of the operating system in the System Under Test [39].

## 3. Data and methodology

In the following sections, we present the data sources and the research methodology used in this study, elucidating how machine learning techniques are applied to model server performance and power consumption.

### 3.1. Data source

This study leverages the SPECpower_ssj2008 database [39] to demonstrate the efficacy of machine learning techniques in server modeling. The SPECPower_ssj2008 database stands as an industry-standard benchmark for volume servers, providing crucial insights into server configuration details, power use, and performance characteristics. This database encompasses servers from leading manufacturers, featuring hardware availability dating back to 2004, as depicted in Fig. 1. As of December 2023, the database includes 949 servers, each with 11 observations at various server workload levels ranging from 0% to 100% and is consistently updated on a quarterly basis.

The publicly accessible SPECPower_ssj2008 database provides a moderate sample size, continuous observations over time, and detailed server characterization information. These attributes make it an valuable resource for employing a machine learning approach in server modeling and generating insightful analyses. However, several biases may limit its applicability: (1) vendors often optimize servers specifically for the SPECpower_ssj2008 benchmark, potentially creating the illusion of overly efficient servers; (2) it focuses exclusively on a Java-based workload benchmark, neglecting other common data center tasks like web hosting, database management, and machine learning training or inference; and (3) tests are performed in controlled environments, disregarding real-world factors such as varying temperatures and dynamic workloads.

Nevertheless, this study emphasizes the feasibility of applying a machine learning approach to server modeling while exploring the impact of server characteristics on power use and efficiency. The proposed framework is designed with flexibility, allowing for updates and expansions as new data or external databases become available.

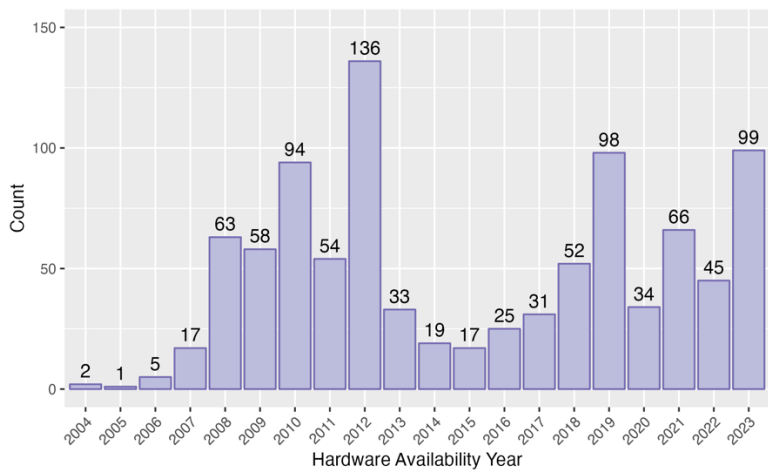

Fig. 1. Distribution of server samples by hardware availability year in the SPECpower_ssj2008 database.

## 3.2. Data preprocessing

Data preprocessing is a pivotal initial step in building machine learning models, as it profoundly influences the capability of machine learning algorithms to extract meaningful information from a dataset. This study's data preprocessing involves two essential steps: (1) data cleaning and unit conversion, and (2) feature selection.

In the first step, data cleaning and unit conversion, we excluded extraneous information, such as hardware vendors and disclosure uniform resource locators (URLs), from the raw data. Additionally, multi-node servers were excluded to concentrate our analysis on the impact of server configuration on power, maximum throughput, and performance-to-power ratio. Textual data were transformed into either numeric or categorical variables using predefined rules from the SPECpower_ssj2008 result file fields [43]. Additionally, we normalized the measurement units of data features to ensure consistency across all observations (Table 1).

The subsequent step, feature selection, was guided by the principle of maintaining model user-friendliness and practical application. Specifically, features were carefully chosen to include only those variables that are typically provided in manufacturers' data sheets or accessible in the public domain. This approach guarantees that model users can rely on readily accessible data for predictive analytics, thereby facilitating informed decision-making. Consequently, the model incorporates features spanning three server configuration categories: central processing unit (CPU), memory, and storage drive. Additionally, server hardware availability date and server workload level were included to monitor server technology development over time and investigate server power proportionality. Collectively, these features offered insights into the technological, temporal, and operational impacts on server power, maximum throughput, and performance-to-power ratio. Table 1 summarizes the feature categories and corresponding server features considered in this study.

Table 1. Sever features for server modeling in this study.

| Category | Sever feature (abbreviations) | Note |
|---|---|---|
| **Central processing unit (CPU)** | Chip count (CC) | CC is the number of chips in the server. |
| | Cores per chip (CPC) | CPC is the number of cores per chip. |
| | Threads per core (TPC) | TPC is the number of hardware threads per core. |
| | Clock frequency (CF) | CF is the nominal clock frequency of the CPU ($MHz$). |
| | L1 data cache size (CS-L1D) | CS-L1D is the CPU's primary data cache, normalized to the average size per core ($\frac{KB}{core}$), it is a numeric variable derived from textual data in the raw dataset. |
| | L1 instruction cache size (CS-L1I) | CS-L1I is the CPU's primary instruction cache, normalized to the average size per core ($\frac{KB}{core}$), it is a numeric variable derived from textual data in the raw dataset. |
| | L2 cache size (CS-L2) | CS-L2 is the CPU's secondary cache (including data and instruction cache), normalized to the average size per core ($\frac{MB}{core}$), it is a numeric variable derived from textual data. |

|  | L3 cache size (CS-L3) | CS-L3 is the CPU's tertiary cache (including data and instruction cache), normalized to the average size per chip ($\frac{MB}{chip}$), it is a numeric variable derived from textual data. |
|---|---|---|
| **Memory** | DIMM count (MMC) | MMC is the number of memory modules in the server, it is a numeric variable derived from textual data. |
|  | DIMM size (MMS) | MMS represents the memory size (in $GB$) per module, it is a numeric variable derived from textual data. |
| **Storage** | Disk drive count (DDC) | DDC is the count of disk drives in the server, it is a numeric variable derived from textual data. |
|  | Disk drive size (DDS) | DDS represents the storage capacity (in $GB$) per disk, it is a numeric variable derived from textual data. |
|  | Disk drive type (DDT) | DDTs include hard disk drive (HDD) and solid-state drive (SSD). It is a categorical variable derived from textual data, which was subjected to one-hot-encoding [44] prior to model training. |
| **Others** | Hardware availability date (HAD) | HAD represents the datetime when all hardware components of the server were available, and it was converted to the Proleptic Gregorian Ordinal format [45] before model training. |
|  | Workload level (L) | $L$ was only used to model $P_L$ and $Perf_L$, not $Th_{max}$. |

### 3.3. Data splitting and modeling pipeline

The preprocessed dataset was subjected to data splitting and a meticulously structured modeling pipeline before resulting in the final server models, as illustrated in Fig. 2.

This study designed two distinct data splitting schemes to address different scopes of server modeling. The first scheme focuses on modeling servers with existing technologies, aligning with the scope of most previous studies. In this scheme, the preprocessed dataset is randomly divided into training, validation, and testing sets, representing 80%, 10%, and 10% of the samples, respectively. Importantly, the dataset is split based on the server index to prevent data leakage [45], ensuring that machine learning algorithms infer server power curves (i.e., server power at different workload levels) solely based on server configuration and hardware availability date. Consequently, the models can reliably predict server power curves without relying on specific power usage data under other workloads from the same server.

The second data splitting scheme is specifically crafted to assess the predictability of power consumption, maximum throughput, and performance-to-power ratio for future servers. This investigation holds substantial importance for forward-looking analysis, a dimension often overlooked in previous studies. To achieve this, we implemented a time-series-based data splitting approach, dividing the preprocessed dataset into training, validation, and testing sets based on hardware availability date (detailed in Appendix B).

The split data were subsequently input into a dedicated modeling pipeline (Fig. 2). This pipeline is crafted to encompass various essential steps, including missing data imputation, feature normalization, model training with hyperparameter tuning, and model selection from models built using candidate machine learning algorithms. Specifically, the k-nearest neighbors (KNN) [46] was employed to impute missing values in the dataset, where the Euclidean distance metric was utilized to identify the nearest neighbors, and the missing values were filled using the average feature values of those neighbors. The standard scaler was employed to normalize the data features by subtracting the mean and scaling to unit variance, a process that can accelerate the model learning process and enhance model performance [47]. A set of candidate machine learning algorithms was selected for model training, with Bayesian optimization [48] applied to fine-tune the hyperparameters of each algorithm. The best server models, targeting power, maximum throughput, and performance-to-power ratio, were ultimately selected based on the model performance metrics on the testing set. A brief introduction to the candidate machine learning algorithms, Bayesian hyperparameter tuning, and model performance metrics can be found in Sections 3.4-3.6. The code for the implemented modeling pipeline is accessible through the link provided in Appendix A.

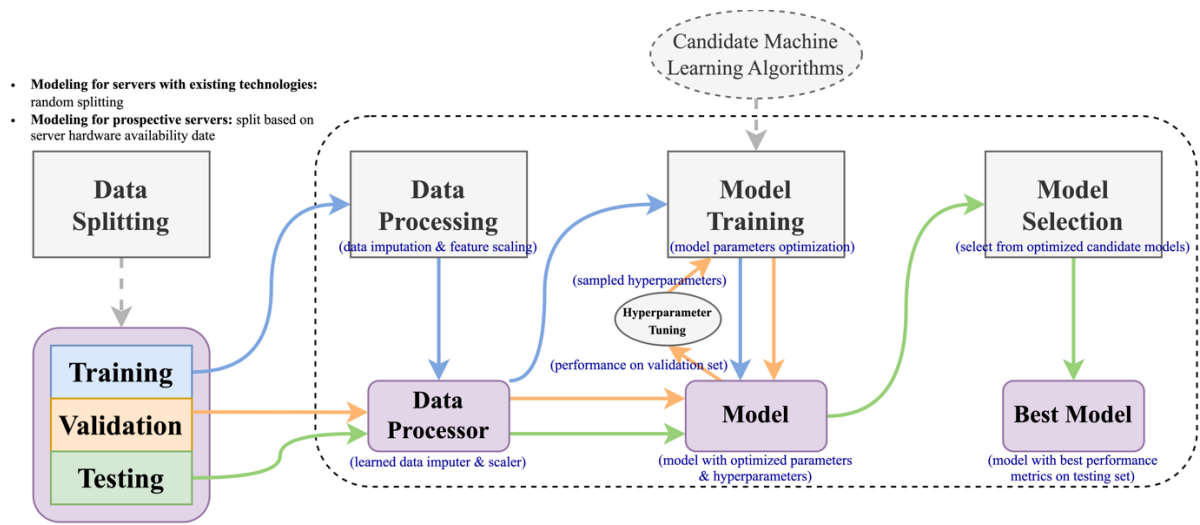

Fig. 2. Flowchart of data splitting and server modeling pipeline.

3.4. Machine learning algorithms

In this study, we developed and compared server models employing a suite of candidate machine learning algorithms, including: (1) elastic-net linear regression (LR), (2) stochastic variational Gaussian process (SVGP), (3) extreme gradient boosting (XGBoost), (4) random forest (RF), and (5) artificial neural network (ANN). A concise overview of these algorithms is presented below, complemented by detailed model and code implementation accessible through the provided link in the Appendix A. Interested readers are directed to pertinent literature for a thorough understanding of these algorithms. The selection of these machine learning algorithms was driven by their varying suitability in addressing nonlinearity and their adaptability to datasets characterized by diverse scales.

The baseline model employed for server modeling was elastic-net LR [49]. This model adeptly merges the regularization techniques of ridge and lasso regression, leading to coefficient shrinkage and variable selection. Such an integrated approach proves advantageous for addressing concerns related to overfitting and multicollinearity within the model. To facilitate a thorough comparative assessment in this research, the elastic-net LR model underwent training on both the original dataset and an augmented set with polynomial features (PFs). These PFs were specifically generated to capture nonlinear associations between independent and dependent variables, with the determination of feature orders accomplished through Bayesian hyperparameter optimization.

SVGP [50] is a recently developed statistical machine learning algorithm designed to alleviate the time complexity associated with training GP models. Its innovation lies in introducing a set of inducing locations to efficiently summarize the entire dataset. SVGP maintains the robust performance of GP models when facing complex function inference and noisy observations. Particularly noteworthy is SVGP's demonstrated improvement in scalability for large datasets, making it a promising choice for our server modeling task, especially given the anticipated growth in the size of training datasets provided by server manufacturers in the future.

XGBoost [51] is an ensemble machine learning algorithm based on decision trees, and it has garnered widespread popularity in the data science community due to its cost-effectiveness during training and its high predictive performance. Employing "gradient boosting" [52], XGBoost addresses model bias by sequentially training trees to fit model residuals, resulting in a robust final-stage model. Users have the flexibility to optimize the model by adjusting tree depth and regularization parameters, effectively reducing variance. Moreover, multiple studies have affirmed the superior performance of XGBoost in modeling nonlinear data [53,54], solidifying its position as another viable candidate algorithm for this study.

RF [55] is another ensemble learning algorithm crafted to enhance model performance through the implementation of bootstrap aggregating. In the training phase, numerous decision trees are concurrently constructed, employing random samples with replacement from the training set and selecting a random subset of features. When predicting unseen samples, RF averages the results provided by the constructed decision trees. Consequently, the RF algorithm contributes to improved model performance by mitigating model variance.

ANNs stand as a powerful and widely adopted model in both industry and academia, finding applications across various domains, including recommender systems, advertising, fraud detection, search ranking, and speech/visual understanding [56]. ANNs are frequently employed for universal function approximation due to their outstanding features such as nonlinear modeling, fault tolerance, and the capability to handle input data in diverse formats [57]. The flexibility of ANNs in data modeling is noteworthy, as they can adapt to different data patterns by adjusting their network structure. Hence, ANN is selected as another candidate algorithm for server modeling in this study.

### 3.5. Bayesian hyperparameter optimization

Hyperparameters dictate the structure and learning process of a machine learning model, making their appropriate setting fundamental for constructing a robust and efficient model. In this study, Bayesian optimization was employed for automatic hyperparameter tuning due to its advantages of reduced time and computational costs, coupled with superior performance when contrasted with conventional methods like manual search, grid search, or random search [48].

The hyperparameter tuning problem using Bayesian optimization can be formulated as Eq. (2), and Table 2 summarizes the Bayesian optimization algorithm utilized in this study.

$$x^* = argmin_x f(x) \quad (2)$$

where $x^*$ represents the optimal set of hyperparameters, $HP$ is the search space defined in Appendix C. Table. C.1, and $f(\cdot)$ denotes a probabilistic surrogate model for the objective function $f$, representing the model's performance on the validation dataset.

Table 2. Bayesian optimization algorithm.

- Step $T = 0$ (initialization):
  Generate initial hyperparameter samples $d^{T=0} = \{(x^{T=0}, y^{T=0}), \cdots\}$, where $x^{T=0}$ is a random hyperparameter sample from Appendix C. Table. C.1, and $y^{T=0}$ is the corresponding model performance on the validation dataset.
- For steps $T = 1$ to $t$:
  (1) Build a GP regression model $f(x)$ based on $d^{T=t-1}$.
  (2) Sample the next point $x^{T=t}$ by maximizing the "expected improvement" acquisition function [58].
  (3) Train the machine learning model based on the new set of hyperparameters $x^{T=t}$ and obtain $y^{T=t}$.

### 3.6. Model performance metrics

The evaluation of server models' performance encompassed metrics such as root mean squared error ($RMSE$), coefficient of determination ($R^2$), mean absolute percentage error ($MAPE$), and mean arctangent absolute percentage error ($MAAPE$), as defined in Eqs. (3) - (6). Notably, the $MAAPE$ metric was selected to address potential numeric issues associated with the $MAPE$ metric, which might yield infinite or undefined values for zero or close-to-zero observations [59]. This choice was made to prevent hindrance in judging model performance. Subsequently, these performance metrics were computed on the training, validation, and testing datasets to quantify the accuracy and generalizability of the models.

$$RMSE = \sqrt{\frac{\sum_{i=1}^{n}(y_i - \hat{y}_i)^2}{n}} \quad (3)$$

$$R^2 = 1 - \frac{\sum_{i=1}^{n}(y_i - \hat{y}_i)^2}{\sum_{i=1}^{n}(y_i - \underline{y})^2} \quad (4)$$

$$MAPE = \frac{1}{n}\sum_{i=1}^{n}\left|\frac{y_i - \hat{y}_i}{y_i}\right| \quad (5)$$

$$MAAPE = \frac{1}{n}\sum_{i=1}^{n} arctan\left(\left|\frac{y_i - \hat{y}_i}{y_i}\right|\right) \quad (6)$$

where $y_i$ is the $i$th observation, $\hat{y}_i$ is the $i$the predicted value, $\underline{y}$ is the sample average, and $n$ is the number of observations.

## 4. Results and discussion

### 4.1. Performance of modeling servers with existing technologies

This section concentrates on evaluating the performance of server models, particularly those pertaining to existing technologies, using the first data splitting scheme introduced in Section 3.3. Fig. 3. provides a summary of the $MAPE$ and $MAAPE$ across various server models concerning target variables and machine learning algorithms. Notably, the XGBoost-based server models consistently outperform other algorithms in predicting server power consumption, maximum throughput, and performance-to-power ratio across all selected performance metrics. Furthermore, the XGBoost-based server models exhibit strong performance on the training, validation, and testing datasets, with less than about 10% on both $MAPE$ and $MAAPE$. This underscores the low bias and high generalization power of the resulting server models.

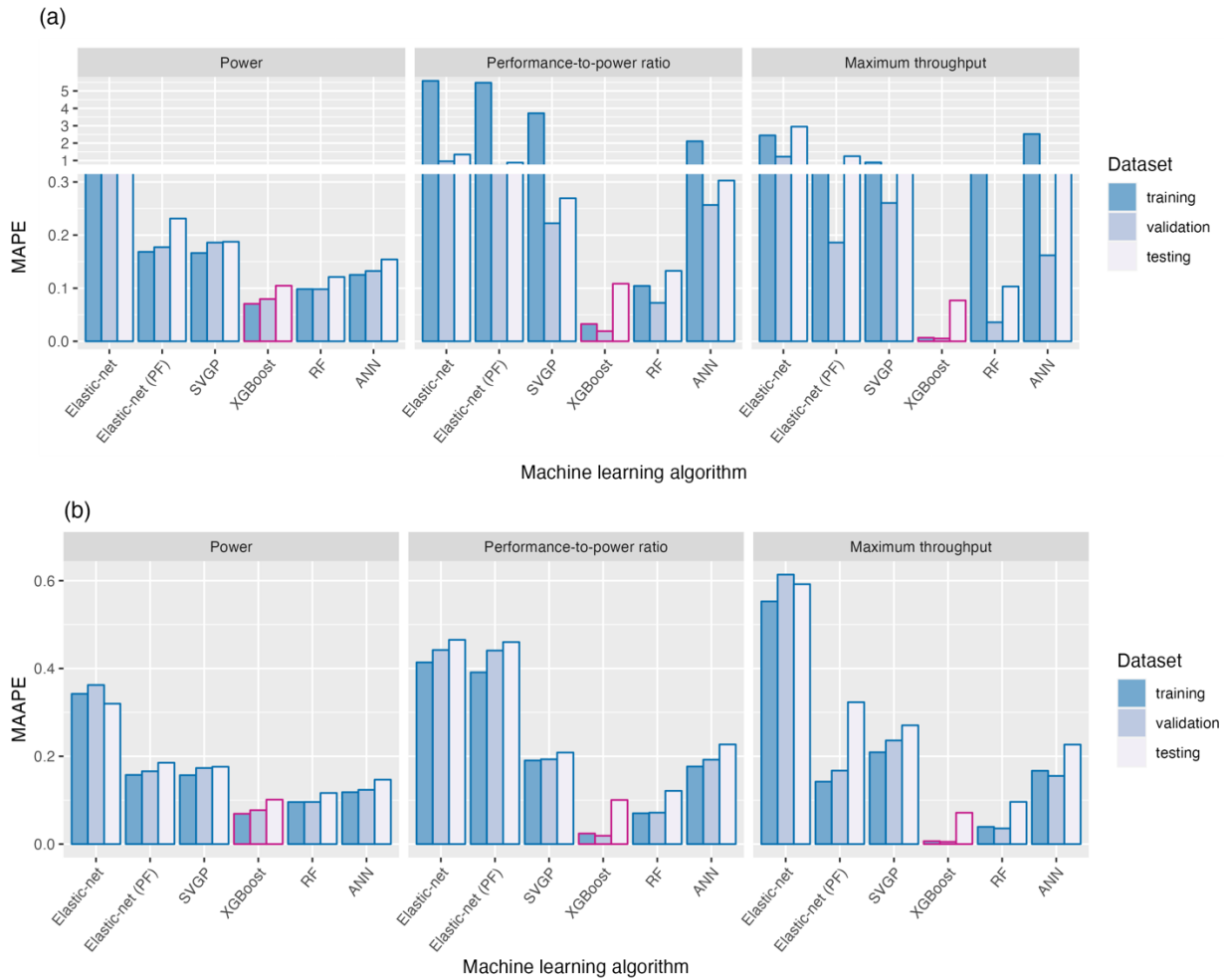

Fig. 3. Performance metrics of trained server models by target variables and machine learning algorithms: (a) $MAPE$, (b) $MAAPE$ (Note: The best server models are indicated by the Magenta color; Performance metrics, including $RMSE$ and $R^2$, are provided in Appendix D).

Fig. 4. illustrates the comparison between observations and predictions obtained from the XGBoost-based server models, highlighting the robust performance and practical utility of our proposed approach. Reflecting on our approach's success, we identify three crucial factors that offer valuable insights for energy analysts and modelers seeking to understand server energy usage and performance through similar machine learning methodologies. Firstly, the SPECPower_ssj2008 database provides an abundant dataset that empowers the model to extract reliable information, even in the presence of data noise. Secondly, the carefully selected server features outlined in Section 3.2 demonstrate their ability to reasonably explain the variances in the target variables. Lastly, the optimized XGBoost-based server models exhibit effectiveness in capturing the nonlinear and complex relationships between the inputs and model targets. It is noteworthy that the model's performance has the potential for improvement with additional data and server features, given the quarterly data updates to the SPECPower_ssj2008 database. However, acquiring additional server features necessitates enhanced reporting from manufacturers. For instance, supplementary details about server configuration, such as PSU efficiency [10] and operating temperature [29], have the potential to elevate the accuracy of future server modeling.

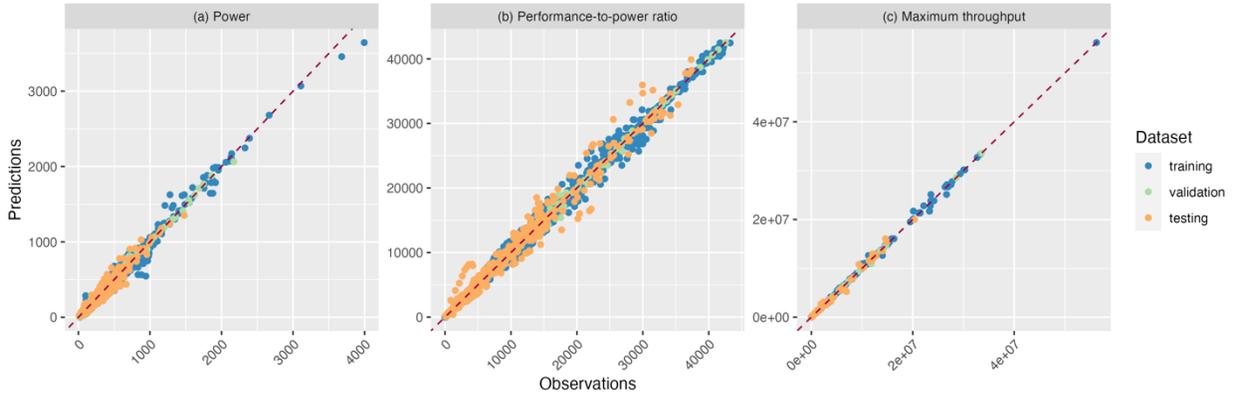

Fig. 4. Comparison of observations vs. predictions by XGBoost-based server models: (a) power consumption ($P_L$, $W$), (b) maximum throughput ($Th_{max}, \frac{ssj\_ops}{s}$), and (c) performance-to-power ratio ($Perf_L, \frac{ssj\_ops}{W}$).

4.2. Important features for modeling servers with existing technologies

To gain additional insights from the trained XGBoost-based models focused on modeling servers with observed technologies, we utilized the permutation feature importance technique [55,60], which identifies and ranks important input variables based on their predictive power. The defined predictive power is measured by the decrease in the model performance metric, such as $R^2$, when a single feature value is randomly shuffled. Fig. 5. illustrates the results of the permutation feature importance generated by the XGBoost models.

The XGBoost models predominantly depend on a few crucial variables for making predictions. It's essential to acknowledge that excluding less important model features may impact model accuracy. Nevertheless, readers have the flexibility to omit lower-ranked features based on their availability and accuracy requirements for specific prediction purposes.

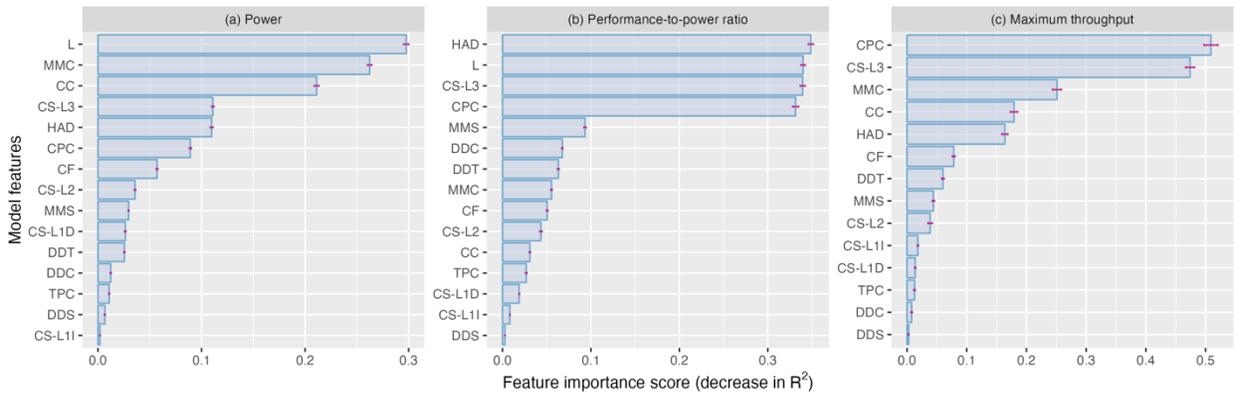

Fig. 5. Permutation feature importance for predicting: (a) power consumption ($P_L$, $W$), (b) maximum throughput ($Th_{max}, \frac{ssj\_ops}{s}$), and (c) performance-to-power ratio ($Perf_L, \frac{ssj\_ops}{W}$) (Note: Refer to Table 1 for the nomenclature).

Furthermore, the analysis highlights the crucial role of temporal and operational factors in server modeling, as indicated by the high rank of HAD and L in Fig. 5. Firstly, the server power at maximum throughput has visibly increased with HAD, though it demonstrates notable variations among servers with different chip counts (Fig. 6 (a)). Secondly, the interplay between HAD and L reveals nuanced interaction effects on server power consumption. This intricacy is portrayed in Fig. 6 (b), illustrating the scaled power of servers operating at different L values. Noteworthy is the observation that, over time, servers tend to become more power-proportional, displaying reduced idle power consumption and a more pronounced linear relationship between server power and L. Thirdly, the influence of HAD and L on server performance-to-power ratio and maximum throughput is evident, as illustrated in Fig. 6 (c) and (d). Both the performance-to-power ratio and maximum throughput show an upward trend with increasing HAD, and a positive correlation is observed between the performance-to-power ratio and L. Specifically, while CPC proves to be the more influential input for predicting maximum throughput, it's noteworthy that the maximum CPC available in servers also rises with HAD, underscoring the significant temporal effects of advancements in server technology. Collectively, the importance of L in server modeling reaffirms the established practice of incorporating it in modeling server power consumption [33,34]. Additionally, the elevated significance of both L and HAD emphasizes the critical role of considering server virtualization technologies and hardware refresh cycles for achieving energy-efficient data center operations [1,4,61]. This insight provides operators with the knowledge to make informed decisions, whether optimizing efficiency through server virtualization or upgrading servers with newer technologies to achieve enhanced energy savings and reduced electricity bills.

Moreover, power consumption, maximum throughput, and performance-to-power ratio are significantly influenced by server configurations. Specifically, MMC and CC are key features for predicting power consumption and maximum throughput, CPC is crucial for predicting performance-to-power ratio and maximum throughput, while CS-L3 is consistently important for predicting power consumption, maximum throughput, and performance-to-power ratio. The importance of these server features aligns with fundamental design principles for computing systems: (1) higher MMC or CC contributes to a larger overall chip area and increased thermal budget, leading to higher maximum throughput and power consumption limit (Fig. 6 (e)) [62]; (2) multicore processors facilitate parallel execution of CPU instructions, enhancing performance-to-power ratio and maximum throughput [63]; and (3) the L3 cache is designed to expedite information flow between main memory and CPU, improving performance-to-power ratio and maximum throughput with additional power usage [64].

Finally, compared to CPU- and memory-specific features, storage-specific features have a less significant role in predicting power consumption, performance-to-power ratio, and maximum throughput. However, we observed a notable impact of DDT on server idle power consumption (Fig. 6 (f)). Servers with SSDs exhibit lower idle power consumption than those with HDDs. While this aligns with previous studies, it's crucial to note the increasing preference for SSDs in newer server generations, where temporal effects also contribute to the observed difference in idle power between the two groups.

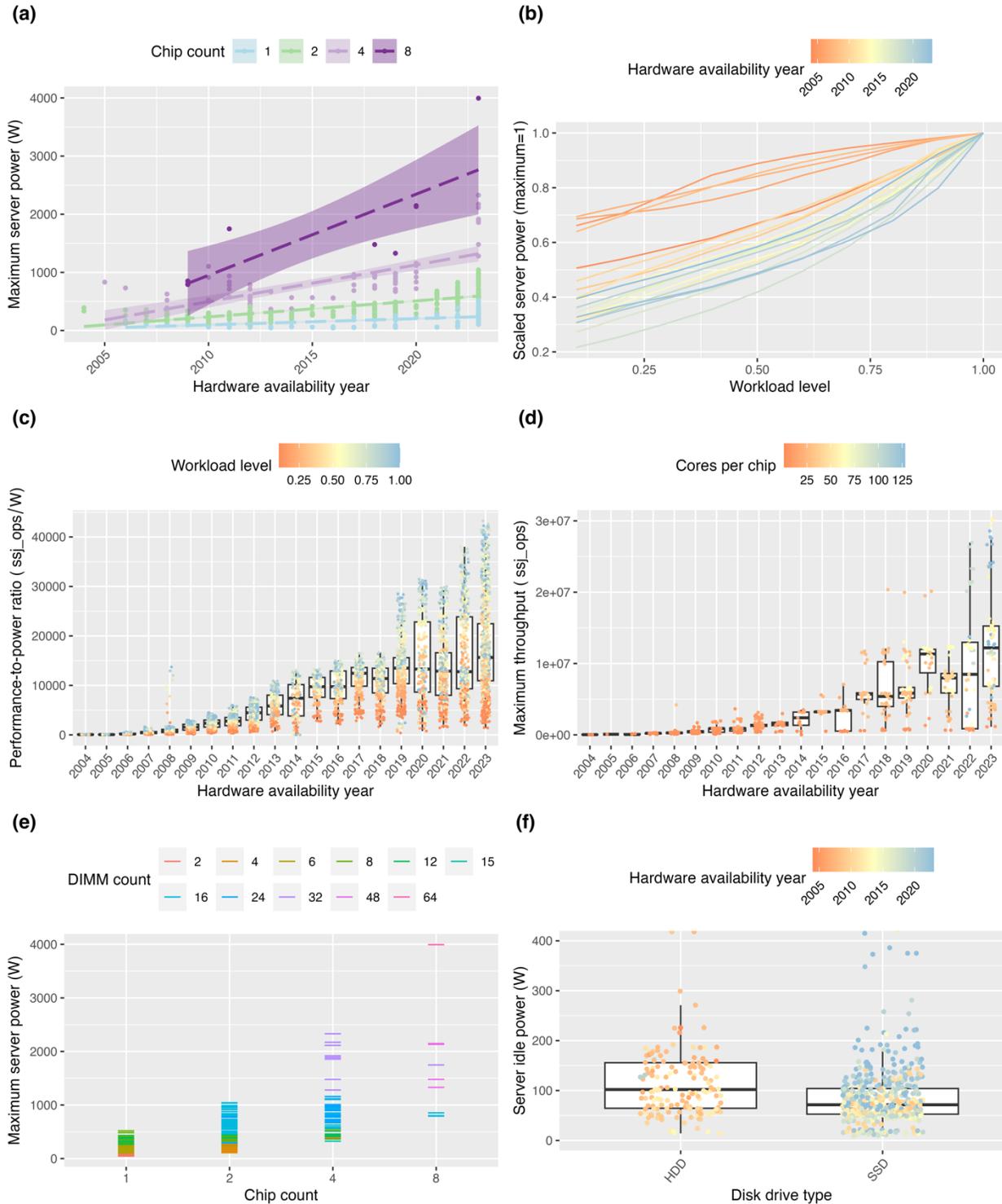

Fig. 6. Relationship between key model features and target variables: (a) server power at maximum throughput vs. hardware availability year and chip count; (b) scaled server power vs. workload level and hardware availability year; (c) performance-to-power ratio vs. hardware availability year and workload level; (d) maximum throughput vs. hardware availability year and cores per chip; (e) server power at maximum throughput vs. DIMM and Chip count; (f) server idle power vs. disk drive type and hardware availability year (Note: The hardware availability year is derived from the aggregation of HAD at the yearly level; the uncertainty bounds in (a) depict the 95% confidence interval for the smoothed trend lines; the servers in (b) represent those with the median dynamic range for each hardware availability year).

### 4.3. The danger of prospective prediction

Fig. 7. illustrates the performance of modeling prospective servers on the testing dataset using the time-series-based data splitting approach (Appendix B), where the $MAAPE$ values are generated based on computer experiments with different baseline years and forecast horizons beyond the baseline years. The performance quality of the server models exhibits a rapid reduction with the extension of forecast horizons, resulting in $MAAPE$ averages exceeding 20% for forward predictions beyond 2 years. In fact, the $R^2$ statistics on the testing datasets quickly drop below zero for forward predictions beyond 2 years (refer to OSM), indicating that the prediction results of the server models are less accurate than using the target mean. In this context, it becomes apparent that employing the machine learning approach based on the SPECpower_ssj2008 database for forward-looking server analysis and predictions may not be suitable. This holds particular significance in the dynamically evolving information and communication technology industry, characterized by unpredictable technological breakthroughs [65]. Hence, energy analysts and policymakers should approach forward-looking analysis with caution, recognizing that predictions based on historical information are susceptible to substantial errors and uncertainties. Nonetheless, further scientific investigations and research, coupled with the acquisition of comprehensive data and insights pertaining to server technologies, are imperative to fortify the robustness and reliability of prospective analyses and predictions.

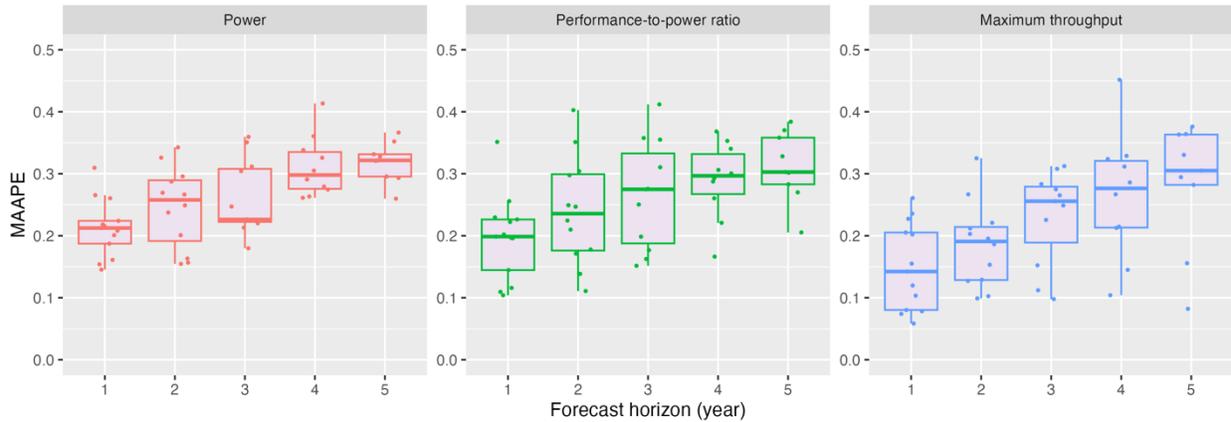

Fig. 7. Modeling performance ($MAAPE$ on testing dataset) of prospective servers across different target variables (Note: The reported results pertain to the performance of the best models. Readers can find performance metric results related to $MAPE$, $RMSE$ and $R^2$ in the OSM).

### 5. Conclusions and future work

This paper investigated the viability of a machine learning approach for modeling server power consumption, performance-to-power ratio, and maximum throughput using the SPECPower_ssj2008 database. The moderately-sized, detailed, and standardized SPECPower_ssj2008 database provides prerequisite data support for server modeling and analysis. The user-friendly features extracted from the database facilitated the development of a systematic machine learning approach, incorporating data imputation, feature scaling, Bayesian hyperparameter optimization, and model selection. The proposed machine learning approach effectively captures intricate relationships between model targets, server configuration, workload level, and hardware availability date, demonstrating consistently low *MAPE* and *MAAPE* values below about 10% when modeling servers with existing technologies. The permutation feature importance method identifies the most predictive parameters in server modeling, emphasizing the significance of technological, operational, and temporal factors. This research further investigated the reliability of prospective server modeling, uncovering significant errors and uncertainties in forward predictions.

The resulting XGBoost models excel in predicting server power consumption, maximum throughput, and performance-to-power ratio across diverse server component specifications and workload levels. This capability is invaluable for optimizing servers in terms of power consumption, cooling budget, power proportionality, throughput, and energy efficiency. Additionally, it holds significant importance for designing and optimizing data centers, covering aspects such as capacity planning, load predictions, cooling arrangements, workload coordination, and operational controls. Furthermore, these models offer clear insights into the relationship between servers' energy consumption, performance, and efficiency with the hardware availability year. This illuminates opportunities for data center energy conservation and cost reductions through optimized hardware refresh cycles and server procurement strategies. Moreover, this study delved into the predictability of the machine learning approach for prospective server modeling, offering complementary insights to previous research. It illuminates the biases and uncertainties inherent in model predictions based on the SPECPower_ssj2008 database. The findings serve as a cautionary note for energy analysts and policymakers engaging in prospective analyses that lean on historical information. Such analyses should be approached with caution to ensure the reliability of the results.

This study can be extended by future work from several aspects. First, while this study relies on the SPECpower_ssj2008 database for model training and validation, external datasets can be leveraged to evaluate the applicability of the resulting model to servers that differ significantly from those in the SPECpower_ssj2008 database. Additionally, the features used in this study are limited to general server specifications, such as CPU, memory, storage, hardware availability dates, and workload levels. Future server models could benefit from integrating additional features encompassing aspects such as operating temperature [29,66], the nature of computing workloads (e.g., artificial intelligence (AI), video streaming, social media) [3,67], the type of server accelerator cards (e.g., graphics processing unit, field-programmable gate array, application-specific integrated circuit, AI accelerators) [67,68], and circuit-level technological details [37]. Incorporating such features can provide a more tailored and flexible modeling framework, addressing specific objectives and leveraging available data for diverse applications. Furthermore, the SPECpower_ssj2008 database is created based on servers operating in controlled environments. Future data gathered from real-world operational settings can serve as a critical supplement to explore factors such as workload variability, cooling strategies (e.g., liquid- or air-cooled servers), and power consumption dynamics under fluctuating conditions. Moreover, upcoming research endeavors should focus on exploring new datasets and methodologies to enhance prospective server modeling. This will contribute to refining the accuracy and reliability of predictions in the evolving landscape of server technologies. Finally, future research endeavors should prioritize the modeling of storage and network devices [1,4,11], offering the potential to expand the scope of this study and provide a comprehensive depiction of electricity usage and energy efficiencies across various categories of information technology equipment within data centers.

## Acknowledgement

The authors express gratitude for the support from Leslie and Mac McQuown, which contributed to part of the research conducted at Northwestern University.

The authors gratefully acknowledge support from the U.S. Department of Energy, Office of Energy Efficiency and Renewable Energy, Industrial Efficiency and Decarbonization Office. Lawrence Berkeley National Laboratory under Contract No. DE-AC02-05CH11231 with the U.S. Department of Energy.

Appendix

Appendix A. Online supplementary materials (OSM)

Supplementary data and code related to this article can be accessed via request.

Appendix B. Time-series-based data splitting

The time-series-based data splitting scheme is tailored for modeling prospective servers. In this approach, the preprocessed data is organized chronologically based on hardware availability date. Subsequently, the dataset is partitioned into training, validation, and testing sets with respect to a chosen baseline year. For data preceding the baseline year, the initial 80% is designated as the training set, while the remaining 20% is reserved for validation to optimize hyperparameters. The resulting optimized models, generated from candidate machine learning algorithms, undergo evaluation on testing sets, which correspond to different years beyond the selected baseline year. This evaluation aids in selecting candidate machine learning models, where the performance of modeling prospective servers is gauged by the best models derived from the model selection process based on the testing data of specific years beyond the baseline year. Notably, this research conducted computer experiments covering various baseline years (2010 to 2022) and different years (1 to 5) beyond the baseline year to comprehensively quantify the performance of modeling prospective servers.

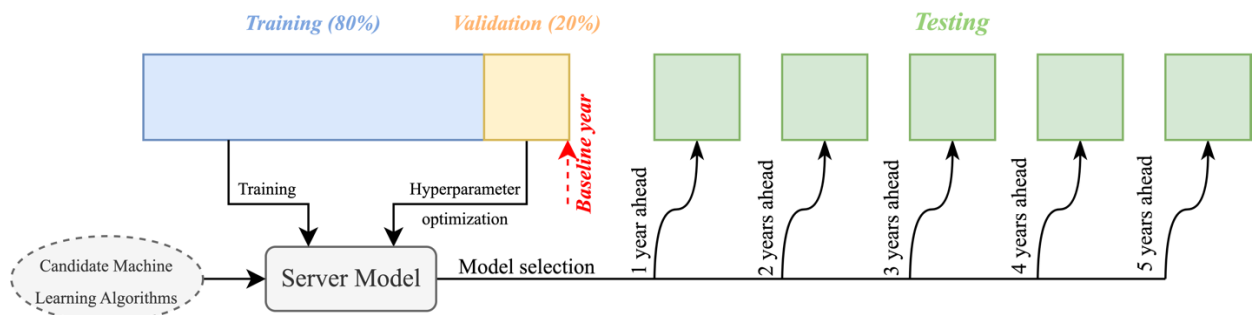

Fig. B.1. Time-series-based data splitting.

Appendix C. Hyperparameters and search space

Table. C.1. Hyperparameters and search space for candidate machine learning models.

| Model | Hyperparameter | Range | Probability distribution |
|---|---|---|---|
| Elastic-net LR | The elastic-net mixing parameter | [0, 1] | Continuous uniform |
| Elastic-net LR (PFs) | The elastic-net mixing parameter | [0, 1] | Continuous uniform |
| | The maximum degree of the polynomial features | [1, 4] | Continuous uniform |
| SVGP | Number of inducing locations | [30, 256] | Integer uniform |

|  | Kernel function | $\{k^{rbf}, k^{M1/2}, k^{M3/2}, k^{M5/2}\}$[1] | Multinomial[2] |
|---|---|---|---|
|  | Learning rate | $[10^{-5}, 10^0]$ | Logarithmic uniform |
| XGBoost | Subsample ratio of columns when constructing each tree | $[0, 1]$ | Continuous uniform |
|  | Subsample ratio of the training instance | $[0, 1]$ | Continuous uniform |
|  | Maximum depth of a tree | $[1, 10]$ | Integer uniform |
|  | Number of boosting rounds | $[1000, 20000]$ | Integer uniform |
|  | L1 regularization term on weights | $[0, 10^3]$ | Continuous uniform |
|  | L2 regularization term on weights | $[0, 10^3]$ | Continuous uniform |
|  | Learning rate | $[10^{-5}, 10^0]$ | Logarithmic uniform |
| RF | Subsample ratio of columns when constructing each tree | $[0, 1]$ | Continuous uniform |
|  | Subsample ratio of columns for each level | $[0, 1]$ | Continuous uniform |
|  | Subsample ratio of columns for each node split | $[0, 1]$ | Continuous uniform |
|  | Maximum depth of a tree | $[1, 10]$ | Integer uniform |
|  | The size of the forest to be trained | $[1000, 2000]$ | Integer uniform |
|  | L1 regularization term on weights | $[0, 10^3]$ | Continuous uniform |
|  | L2 regularization term on weights | $[0, 10^3]$ | Continuous uniform |
|  | Learning rate | $[10^{-5}, 10^0]$ | Logarithmic uniform |
| ANN[3] | Hidden layers | $[0, 5]$ | Integer uniform |
|  | Hidden nodes | $[10, 200]$ | Integer uniform |
|  | Dropout rate | $[0.05, 0.3]$ | Continuous uniform |
|  | Learning rate | $[10^{-5}, 10^0]$ | Logarithmic uniform |

Note: [1] $k^{rbf}$, $k^{M1/2}$, $k^{M3/2}$, and $k^{M5/2}$ denotes radial-basis, Matérn $\frac{1}{2}$, Matérn $\frac{3}{2}$, and Matérn $\frac{5}{2}$ kernel, respectively (each kernel has an additional additive white noise kernel to account for the white noise contribution) [50,69]; [2] Each event has an equal probability of occurring; [3] In our ANN implementation, the number of hidden nodes and dropout rate are consistent in each hidden layer, and rectified linear unit (ReLu) activation function [70] was used.

Appendix D. $R^2$ and $RMSE$ of modeling servers with observed technologies

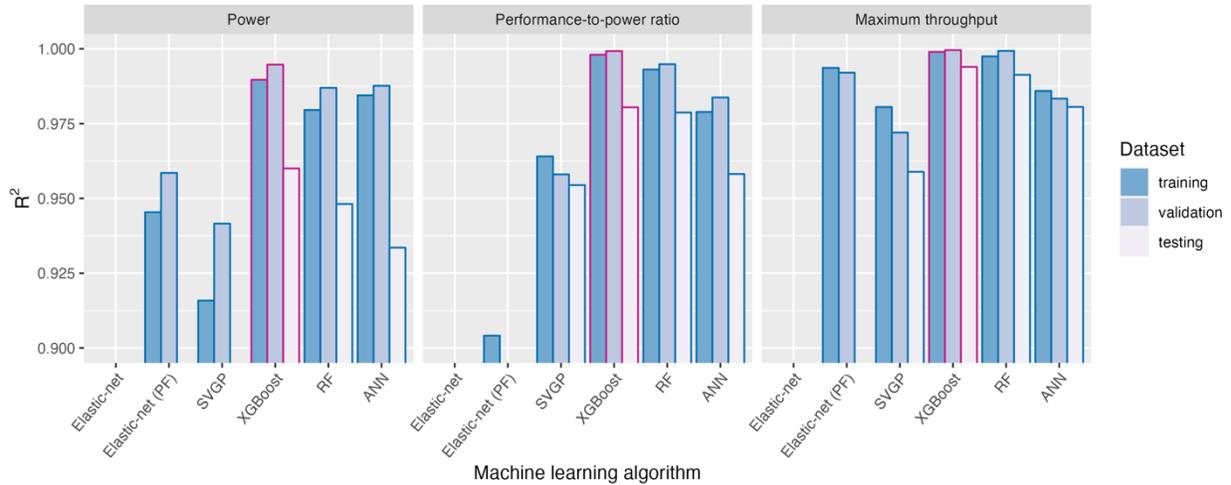

Fig. D.1. $R^2$ of trained server models by target variables and machine learning algorithms (Note: The best server models are indicated by the Magenta color; The results do not show models with negative $R^2$ values).

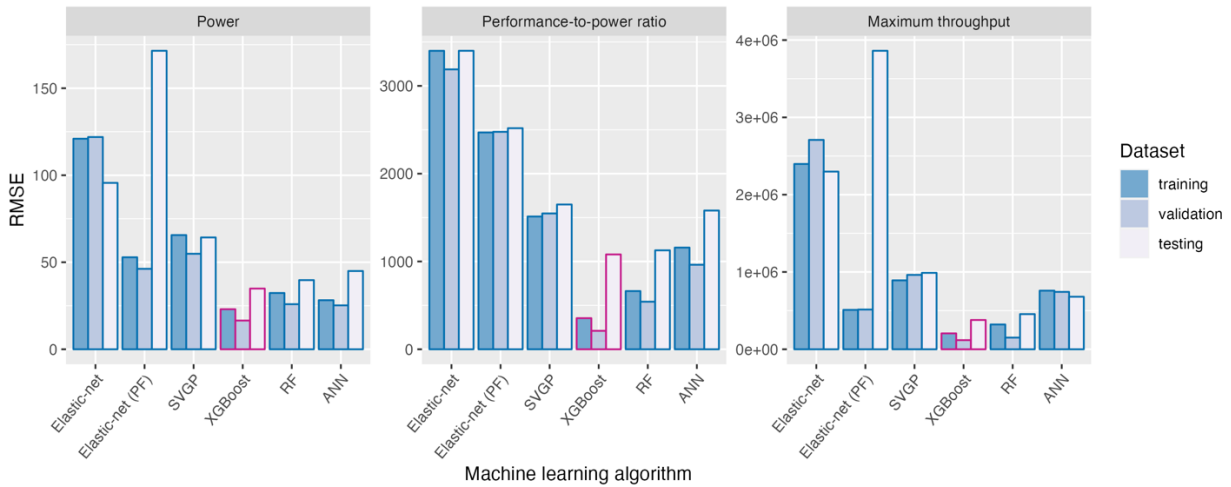

Fig. D.2. $RMSE$ of trained server models by target variables and machine learning algorithms (Note: The best server models are indicated by the Magenta color).